\documentclass[10pt,twocolumn,letterpaper]{article}

\usepackage{wacv}
\usepackage{times}
\usepackage{epsfig}
\usepackage{graphicx}
\usepackage{amsmath}
\usepackage{amssymb}
\usepackage{subcaption}
\usepackage{booktabs}
\usepackage{multirow}
\usepackage{diagbox}
\usepackage{animate}

\wacvfinalcopy 

\def\wacvPaperID{156} 

\ifwacvfinal\fi
\begin{document}

\title{Deep Two-Stream Video Inference for Human Body Pose and Shape Estimation}
\author{Ziwen Li\textsuperscript{1}, Bo Xu\textsuperscript{1}, Han Huang\textsuperscript{1}, Cheng Lu\textsuperscript{2} and Yandong Guo\textsuperscript{1,*}\\
\textsuperscript{1}OPPO Research Institute, \textsuperscript{2}Xmotors\\
{\tt\small yandong.guo@live.com}
}
\maketitle
\ifwacvfinal\thispagestyle{empty}\fi
\pagestyle{empty}  
\thispagestyle{empty} 
\begin{abstract}
Several video-based 3D pose and shape estimation algorithms have been proposed to resolve the temporal inconsistency of single-image-based methods. However it still remains challenging to have stable and accurate reconstruction. In this paper, we propose a new framework Deep Two-Stream Video Inference for Human Body Pose and Shape Estimation (DTS-VIBE), to generate 3D human pose and mesh from RGB videos. We reformulate the task as a multi-modality problem that fuses RGB and optical flow for more reliable estimation. In order to fully utilize both sensory modalities (RGB or optical flow), we train a two-stream temporal network based on transformer to predict SMPL parameters. The supplementary modality, optical flow, helps to maintain temporal consistency by leveraging motion knowledge between two consecutive frames. The proposed algorithm is extensively evaluated on the Human3.6 and 3DPW datasets. The experimental results show that it outperforms other state-of-the-art methods by a significant margin.
\end{abstract}


\section{Introduction}
Considerable amount of research has been done on the 3D human pose and shape estimation from a single RGB image~\cite{kocabas2020vibe,varol2018bodynet,kolotouros2019learning,kanazawa2019learning}. More recently, some methods try to improve 3D human reconstruction by exploiting temporal information from monocular video~\cite{kocabas2020vibe,kanazawa2019learning}. However, those methods still struggle to reconstruct accurate 3D human body when there is complex human joint movement or severe occlusion, largely because of limited sensory modality and training data.



\begin{figure}[t]
\centering
    \animategraphics[width=8cm,height=4.5cm, autoplay, loop]{15}{img_}{00093}{00159}

\caption{DTS-VIBE learns the complementation of multiply sensory modalities for 3D human mesh reconstruction. Given a RGB video (a), we first estimate its optical flow (b) by auto encoder and predict the human mesh (c) by our two-stream temporal encoder-to-decoder network based on transformer. \emph{We suggest readers view this animated figure by Adobe Reader.}}
\label{architecture}
\end{figure}



To address this, we revisit the 3D human reconstruction from video with the following two beliefs. First, we argue that RGB feature alone is insufficient to interpret the high degree of freedom of human behavior, and additional sensory information is needed. Second, we argue that a much stronger temporal network should be introduced into 3D human reconstruction, considering the diverse and complex human motion.  

Consequently, we design our reconstruction pipeline with careful reconsideration in the following three ways. \par

First, we propose a new two-stream architecture named {\bf D}eep {\bf T}wo-{\bf S}tream {\bf V}ideo {\bf I}nference for Human {\bf B}ody Pose and Shape {\bf E}stimation (DTS-VIBE), which allows multi-modality fusion for 3D human reconstruction. With this new architecture, we can supplement other sensory modalities, including but not limited to depth, optical flow or other motion cues, to extend RGB feature space for better 3D human pose and shape estimation. To the best of our knowledge, this is the first time such multi-modal architecture is introduced into this area to simultaneously compensate motion instability and improve temporal consistency. This architecture is highly modularized and each of its component is interchangeable, so it can easily accommodate other modalities and their own optimal seq2seq encoder/decoder. \par 
Second, we select optical flow among those candidate modalities to fit the above architecture. The optical flow has proven to be highly effective to understand human behaviour~\cite{simonyan2014two} because it can, intuitively, help the two-stream network to bridge each two adjacent frames by understanding the motion. It is worth noting that our optical flow input does not require actual physical sensor such as event camera, which causes extra cost on data collection, labelling and synchronization with RGB image. Instead, we estimate the optical flow from RGB image sequence as a virtual sensory modality to provide an applicable and inexpensive solution. In other words, optical flow is directly estimated from RGB images without any additional labels.\par

Third, to correlate video frames in a way that best serves the pose and shape estimation, we build our temporal encoder based on transformer~\cite{vaswani2017attention} instead of Gated Recurrent Units (GRU)~\cite{cho2014learning} which is used as the temporal network in the state-of-the-art method \emph{VIBE}~\cite{kocabas2020vibe}. Despite being widely applied on various temporal tasks, GRU has its shortcomings that the temporal information is inevitably lost during the recursion. For example, when human body is partially occluded in the leading frames of a sequence, GRU cannot effectively provide reliable temporal information to those frames because referring to prior frames are impossible. However, transformer can alleviate this situation by applying global multi-head attention, particularly to the latter frames that are highly correlated to those leading frames in this case, to better estimate shapes and poses. Similarly, when human body is occluded in the middle of a sequence, transformer can combine short-term and long-term attentions simultaneously to better infer and regularize the motion in the middle. To fully exploit the proposed transformer network, we introduce a new loss named $L_{flow}$ which is inspired by \cite{dong2018supervision}. The loss is designed so that optical flow information is used to regularize the estimated pose by enforcing the trajectory of certain motion. We experimentally show that the combination of transformer network and flow loss significantly improves both accuracy and stability.
\par
Overall, we employ dual convolutional neural networks (CNN) to extract two feature streams from RGB image/video sequence and its optical flows. The extracted image feature streams are fed into a transformer-based temporal encoder, and then combined with the corresponding optical flow streams by late fusion. The fused feature will be finally fed into a transformer-based regressor~\cite{vaswani2017attention}. Then, we follow~\cite{kocabas2020vibe} to add a discriminator that tries to distinguish between the regressed body poses and samples from the AMASS~\cite{kanazawa2019learning} dataset, which can provide a real/fake label for each sequence.

\begin{figure*}[t]
\centering
    \includegraphics[width=1.0\linewidth]{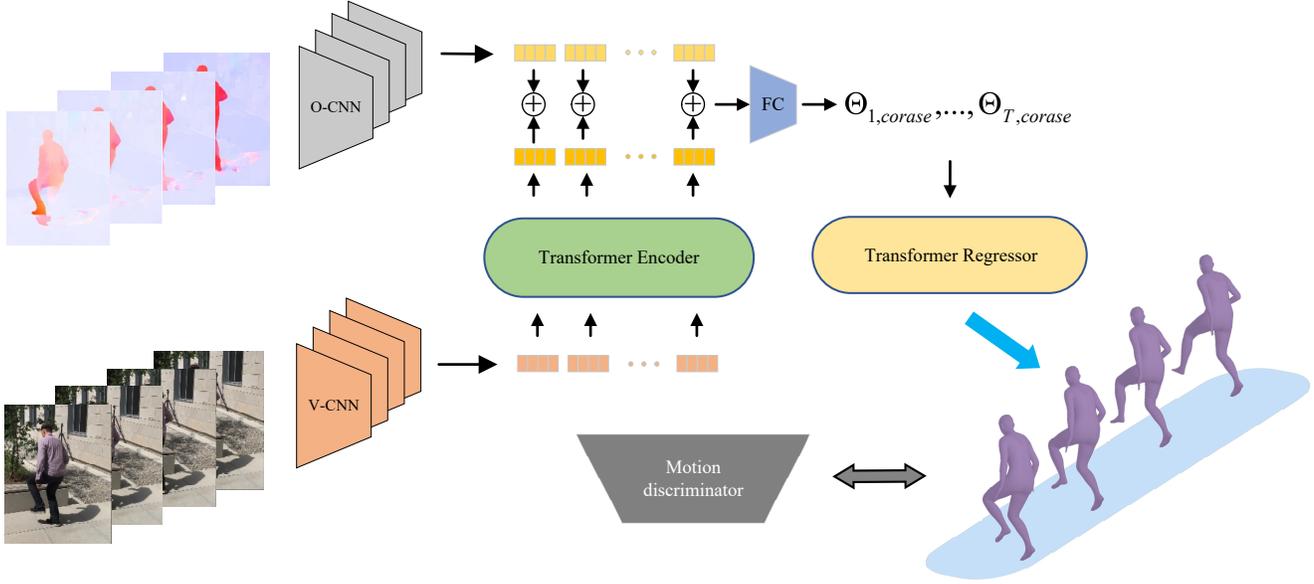}

\caption{\textbf{Architecture of the DTS-VIBE}. The sequence of images and its corresponding flow are sent to the CNN to extract features. The transformer encoder extract temporal features and adds it with the flow feature. A fully connected layer is used to predict coarse SMPL parameters, which is processed by transformer regressor, providing final SMPL parameters. The whole network is trained with a discriminator similar to VIBE.}
\label{architecture}
\end{figure*}
To justify our solutions, we conduct extensive experiments on multiple datasets. The experimental results show that our proposed method surpasses all state-of-the-art video-based and single-image-based approaches. Overall, the contributions of this paper are: 
\begin{itemize}
    \item To our best knowledge, this is the first end-to-end two-stream architecture which allows multi-modality fusion for video 3D human reconstruction.
    \item We introduce virtual optical flow which can supplement the corresponding motion information for the RGB domain to predict more accurate pose and shape.  
    \item We propose a transformer-based temporal network to establish more robust temporal correlations and a transformer-based SMPL regressor for better body shape parameterization. 
    \item We introduce a flow loss to regularize the predicted keypoints and reduce the acceleration error.
\end{itemize}

\section{Related works}
\subsection{Single image-based human reconstruction}
Parametric human models~\cite{anguelov2005scape,loper2015smpl,pavlakos2019expressive} are widely used in many prior studies. Pre-trained parametric models such as SMPL~\cite{loper2015smpl}, STAR~\cite{osman2020star}, SMPL-X~\cite{pavlakos2019expressive} are able to recover a human mesh with a few coefficients. Bogo\etal~\cite{bogo2016keep} propose SMPLify, which first uses CNN to detect the 2D keypoints from a given image, then fits SMPL model to the predicted keypoints. Because of the high cost of 3D human data collection and annotation, people tend to use projected 2D keypoints loss as the less ideal supervision. HMR~\cite{kanazawa2018end} directly regresses SMPL parameters given an input image and projects 3D joints to 2D using the estimated camera parameters to match the ground truth 2D keypoints. It also proposes a discriminator network to distinguish the predicted and ground truth parameters, generating a more plausible mesh. Kolotouros\etal~\cite{kolotouros2019learning} proposed SPIN that interatively fits the SMPL model to 2D joint and use the current estimation to supervise the network in the next iteration. \par 
Meanwhile, some methods~\cite{pavlakos2019expressive,yao2019densebody} also use intermediate output to help reconstruct human mesh. Pavlako\etal~\cite{pavlakos2019expressive} first generates 2D joint heatmap and mask, then combines them with input image to predict SMPL parameters. Pengfei\etal~\cite{yao2019densebody} aligns 3D mesh to 2D image and use UV map to represent the 3D human mesh and uses encoder-decoder model to predict the UV map. Besides, there are plenty of other approaches which directly regress 3D human mesh from a given still image~\cite{moon2020i2l,varol2018bodynet,kolotouros2019convolutional,choi2020pose2mesh,saito2019pifu}. Gyeongsik\etal~\cite{moon2020i2l} proposed an Image-to-Lixel network that directly predict three 1D-heatmaps of xyz coordinates of the human mesh veitices. Varol\etal~\cite{varol2018bodynet} proposes BodyNet, which estimates voxels of human shape in the 3D volumetric space.
\subsection{Video-based 3D human reconstruction}
Despite the progress of 3D human pose and shape estimation from single image, there are some video based methods~\cite{kanazawa2019learning,kocabas2020vibe,luo20203d,dabral2018learning,hossain2018exploiting,mehta2018single,pavllo20193d} which take advantage of temporal information and subsequently achieve impressive outcomes. Hossain\etal\cite{hossain2018exploiting} propose a LSTM model to predict a squence of 3D joints from given 2D joints. Pavllo\etal\cite{pavllo20193d} use a fully-convolutional network to process a sequence of 2D joints to estimate its 3D location, and these predicted 3D joints are used to self-supervise the network on the original 2D joints. \cite{arnab2019exploiting} first predict 2D joints and SMPL parameters for each frame, then jointly optimize them to reduce the error. Kanazawa~\etal~\cite{kanazawa2019learning} propose HMMR, which uses 1D CNN as temporal encoder to find features from sequence of images. It predicts 3D poses not only for the target frame but also past and future frames. Such strategy guarantees the smooth estimation. VIBE~\cite{kocabas2020vibe} utilize GRU to encode features from single image into temporal feature and regress SMPL parameters. It also introduces a motion discriminator to guide the generator and encourage it to predict more reasonable poses when compared to the labels in auxiliary dataset. All of them prove to be great success, yet they still lack of temporal consistency in certain degree. In this regard, MEVA~\cite{luo20203d} feed a sequence of human motion into an auto-encoder framework to first get coarse motion with VME, then refine it with MRR. Its network with encoder-decoder architecture learns the motion of human pose from AMASS\cite{mahmood2019amass}, which helps to smooth pose estimation at inference.
\subsection{Two Stream}
Two stream architecture is widely adopted in video action recognition~\cite{simonyan2014two,christoph2016spatiotemporal,lan2017deep,diba2017deep}. The feature obtained from optical flow is intuitively helpful when trying to learn the motion of people. Simonyan\cite{simonyan2014two} propose to extract spatial feature from images and extract temporal feature from dense optimal flow. The two types of features are combined by late fusion and sent to the classification module. Meanwhile, optical flow is also used to help 3D reconstruction. \cite{doersch2019sim2real} combine flow and 2D keypoints to predict poses over the input video and achieve decent performance under occluded scenarios. Similarly, our approach leverages the optical flow to enhance the temporal feature.

\section{Architecture}
In this section, we describe the deep two-stream video inference network for human body pose and shape estimation (DTS-VIBE). As summarized in Figure~\ref{architecture}, the DTS-VIBE model consists of two-stream encoder, transformer-based regressor and motion discriminator. 

\subsection{Two-stream encoder}

The intuition behind using a two-stream encoder to process RGB and optical flow separately is that optical flow can provide additional sensory information of human motion. This is particularly useful for difficult occasions when human is severely occluded or the pose of human is irregular. Optical flow information can also help to maintain the smoothness of human prediction.  

Given a sequence of continuous RGB frames $v_1,...,v_T$, we first group every two consecutive video frames to obtain estimated optical flow $o_1,...,o_T$ using an auto encoder~\cite{Liu_2019_CVPR}. The RGB stream $v_1,...,v_T$ and optical-flow stream $o_1,...,o_T$ are fed into a dual CNNs (V-CNN and O-CNN), which outputs two-stream feature vectors $f^{v}_i \in \mathbb{R}^{2048}$ and $f^{o}_i \in \mathbb{R}^{2048}$, $i \in (1, T)$. We adopt ResNet-50~\cite{christoph2016spatiotemporal} as the backbone network of V-CNN and O-CNN, followed by the fully connected layers to reduce the dimensions from 2048 to 512. Instead of using Gated Recurrent Units (GRU), which demonstrates decent performance on video-based 3D human reconstruction~\cite{kocabas2020vibe}, our temporal network is instead built on transformer~\cite{vaswani2017attention}. Although GRU is widely applied in various temporal tasks, it shows disadvantages that the temporal information is inevitably lost during the recursion. Transformer can alleviate this situation, because it directly works on every feature simultaneously with global multihead attention. 

\begin{figure}[t]
\centering
    \includegraphics[width=0.40\linewidth]{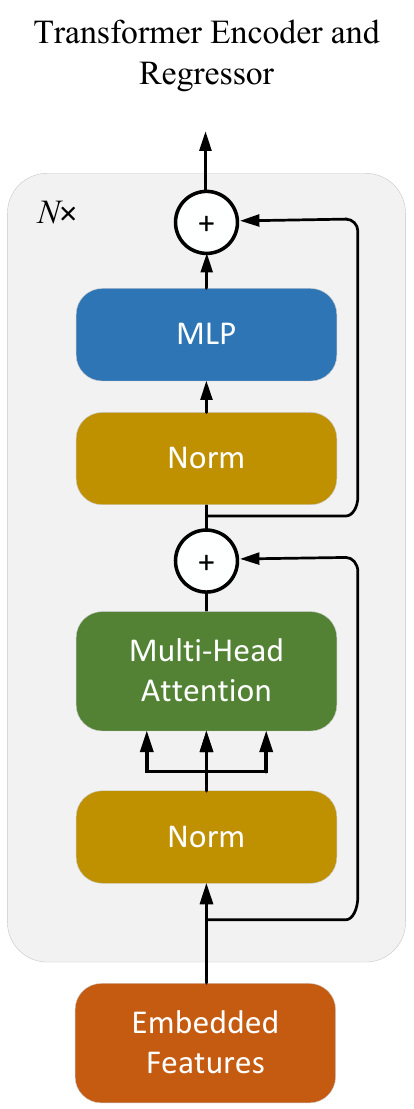}

\caption{\textbf{Architecture of Transformer.} Our temporal encoder use the same transformer structure but with different number of layers or heads.}
\label{transformer}
\end{figure}

The sequence of RGB feature vectors $f^{v}_i \in \mathbb{R}^{512}$ are fed into our transformer~\cite{vaswani2017attention}-based temporal encoder that yields latent RGB feature vectors $g^{v}_i \in \mathbb{R}^{512}$. Then the optical-flow feature vectors $f^{o}_i \in \mathbb{R}^{512}$ are added on $g^{v}_i$ and sent to a transformer-based regressor~\cite{kolotouros2019learning} to predict the per-frame SMPL parameters and corresponding camera parameters. 
Our transformer-based temporal encoder consists of $N$ layers, each of which contains two sub-layers: a multi-head attention layer and a feed forward layer. As shown in Figure~\ref{transformer}, given a sequence of input features $x_{1},...,x_{T}$, the multi-head attention layer first gets the \textit{query} (Q), \textit{key}(K), and \textit{value}(V) by using fully connected layer:

\begin{equation} Q = W^Q X, K = W^K X, V = W^V X\end{equation}

Then the output is computed as 
\begin{equation}\ a = softmax(\frac{QK^T}{\sqrt{d}})V,\end{equation}
where $d$ is the dimension of key. The feed forward layer consists of 2 fully connected layer and dropout.
The output of transformer is then added with the output of O-CNN, which contains the temporal information between each two frames, followed by a fully connected layer to reduce feature dimension from 512 to 157. The transformer regressor takes the low-dimension feature as input and outputs accurate SMPL parameters $\hat{\Theta}_{1},...,\hat{\Theta}_T$
The overall loss of our module is:
\begin{equation}
    L_{total} = \lambda_{1} L_{3D}+\lambda_{2} L_{2D}+ \lambda_{3} L_{smpl}+ \lambda_{4} L_{adv}
\label{overall_loss}    
\end{equation}

where $\lambda_{1}$,$\lambda_{2}$,$\lambda_{3}$,$\lambda_{4}$ are weighted coefficients for each loss.
Specifically, $L_{3D}$ is the $L_2$ distance between ground truth 3D joints location $X_{i,3D}$ and 3D joints location $\hat{X}_{i,3D}$ from predicted SMPL model:
\begin{equation}\ L_{3D} = \sum_{i=1}^{T}||X_{i,3D} - \hat{X}_{i,3D}||_2^2 \end{equation}
where $L_{2D}$ is the $L_{2}$ distance between ground truth 2D joints location and the weak-perspective projection of predicted 3D joints using corresponding camera parameters:
\begin{equation}\ L_{2D} = \sum_{i=1}^{T}||X_{i,2D} - \hat{X}_{i,2D}||_2^2 \end{equation}
where $L_{smpl}$ is the L2 distance between ground truth and predicted SMPL parameters:

\begin{equation}\ L_{smpl} = \sum_{i=1}^{T}||\theta_{i} - \hat{\theta}_{i}||_2^2 + ||\beta_{i} - \hat{\beta}_{i}||_2^2 \end{equation}

\subsection{Motion discriminator and flow supervision}
{\bf Motion discriminator.} Follow~\cite{kocabas2020vibe}, we use a motion discriminator $D_{M}$ to distinguish the predicted SMPL parameters $\theta$(fake) and the real SMPL parameters $\hat{\theta}$(real) from AMASS~\cite{mahmood2019amass}. Motion discriminator helps to produce more feasible real world poses that are aligned with 2D joint locations:

\begin{equation}\ L_{adv} = E_{\Theta\sim p_G}[(D_M(\hat{\Theta})-1)^2] \end{equation}

The objective for the discriminator is:
\begin{equation}
    L_{D} = E_{\Theta\sim p_R}[(D_M(\Theta)-1)^2 + E_{\Theta\sim p_G}[D_M(\hat{\Theta})^2]
\end{equation}
where $p_G$ is a generated motion sequence and $p_R$ is a real motion sequence from the AMASS dataset. 

\begin{table*}[t]
\centerline{
\setlength{\tabcolsep}{1mm}{
\scalebox{0.95}{
\begin{tabular}{l l|c c c c|ccc|ccc}

\toprule

\multicolumn{2}{c|}{\multirow{2}*{\bf Method}}& \multicolumn{4}{c}{3DPW}& \multicolumn{3}{c}{MPI-INF-3DHP} &  \multicolumn{3}{c}{Human3.6}\\
\cline{3-12}
\multicolumn{2}{c|}{~}& PA-MPJPE↓& MPJPE↓& PVE↓& Accel↓ & PA-MPJPE↓& MPJPE↓& Accel↓ & PA-MPJPE↓& MPJPE↓& Accel↓\\

\midrule
\multirow{5}*{\rotatebox{90}{\bf Single image}}\qquad & HMR & 76.7&130.0&-&37.4&89.8&124.2&-&56.8&88.0&-\\
~ & GraphCMR & 70.2& -& -& - & -& -& - & 50.1& -& -\\
~ & SPIN & 59.2 & 96.9 & 116.4 & 29.8 & 67.5 & 105.2 & - & 41.1 & - & 18.3\\
~ & I2L-MeshNet & 57.7 & 93.2 & 110.1 & 30.9 & - & - & - & 41.1 & 55.7 & 13.4\\
~ & Pose2Mesh & 58.3 & 88.9 & 106.3 & 22.6 & - & - & - & 46.3 & 64.9 & 23.9\\
\hline
\multirow{6}*{\rotatebox{90}{\bf Video}}\qquad & HMMR & 72.6 & 116.5 & 139.3 & 15.2 & - & - & - & 56.9 & - & -\\
~ & Sun et al. & 69.5 & - & - & - & - & - & - & 42.4 & 59.1\\
~ & SPIN & 59.2 & 96.9 & 116.4 & 29.8 & 67.5 & 105.2 & - & 41.1 & - & 18.3\\
~ & MEVA & 54.7 & 86.9 & - & 11.6 & 65.4 & 96.4 & \textbf{11.1} & 53.2  & 76.0 & 15.3\\
\cline{2-12}
~ & VIBE & 51.9 & 82.9 & 99.1 & 23.4 & 64.6 & 96.6 & 27.3 & 41.4 & 65.6 & -\\
\cline{2-12}
~ & DST-VIBE(Ours) & \textbf{50.3} & \textbf{76.7} & \textbf{93.5} & \textbf{11.0} & \textbf{62.2} & \textbf{93.4} & 11.9 & \textbf{39.3} & \textbf{60.5} & \textbf{5.0}\\

\bottomrule
\end{tabular}}}}
\caption{\textbf{Evaluation of state-of-the-art methods on 3DPW, MPI-INF-3DHP, and Human3.6M datasets.} Models of VIBE,MEVA and DST-VIBE use 3DPW train set. And all methods except MEVA use Human3.6 for training. So the most fair comparison is between VIBE and ours. DST-VIBE achieves state-of-the-art results with almost all the metrics.}\label{tab:aStrangeTable}
\end{table*}

{\bf Flow supervision.} Inspired by \cite{dong2018supervision}, we introduce a new flow loss to refine our DTS-VIBE network. Since optical flow represents the motion of two adjacent frames, the 2D joints location should move following its corresponding flow. So we use our flow sequence $o_1,...,o_T$ to supervise the movement of 2D joints location.We back trace the previous frame 2d joint position $X^f_{i-1,2D}$ using optical flow $o_i$ and current 2d joint position $X_{i,2D}$, the flow loss is calculated as $L_2$ distance between backward 2D joints and original 2D joints:
\begin{equation}\ L_{flow} = \sum_{i=2}^{T}||X_{i-1,2D} - X_{i-1,2D}^f||_2^2 \end{equation}Note that we do not add flow loss at the beginning of the training, we only use it as supervision during refinement.\par
Among the dual convolutional network we use, the V-CNN is pre-trained on frame-based pose and shape estimation task~\cite{kolotouros2019learning}, while the O-CNN is pre-trained on the ImageNet~\cite{russakovsky2015imagenet} dataset. Similar to~\cite{kocabas2020vibe}, we choose sequence length $T = 16$, which also leads to the best results compared with other alternative ones $i.e.$ 8, 32, 64 and 128. For the transformer-based temporal encoder, we choose the number of layers to be 6 and number of head to be 8. The transformer-based regressor has 3 layers with 4 heads. We set the four loss term $\lambda_1, \lambda_2, \lambda_3, \lambda_4$ in Eq.~\ref{overall_loss} to 300, 200, 120 and 60 respectively. We use Adam optimizer~\cite{kingma2014adam} with learning rate 5 $\times 10^{-5}$ for the generator and 5 $\times 10^{-4}$ for the motion discriminator. During the refinement with the flow loss $L_{flow}$ as supervision, we remove the discriminator and use Adam optimizer~\cite{kingma2014adam} with learning rate 1 $\times 10^{-5}$.

\begin{figure*}[t]
\centering
    \includegraphics[width=1.0\linewidth]{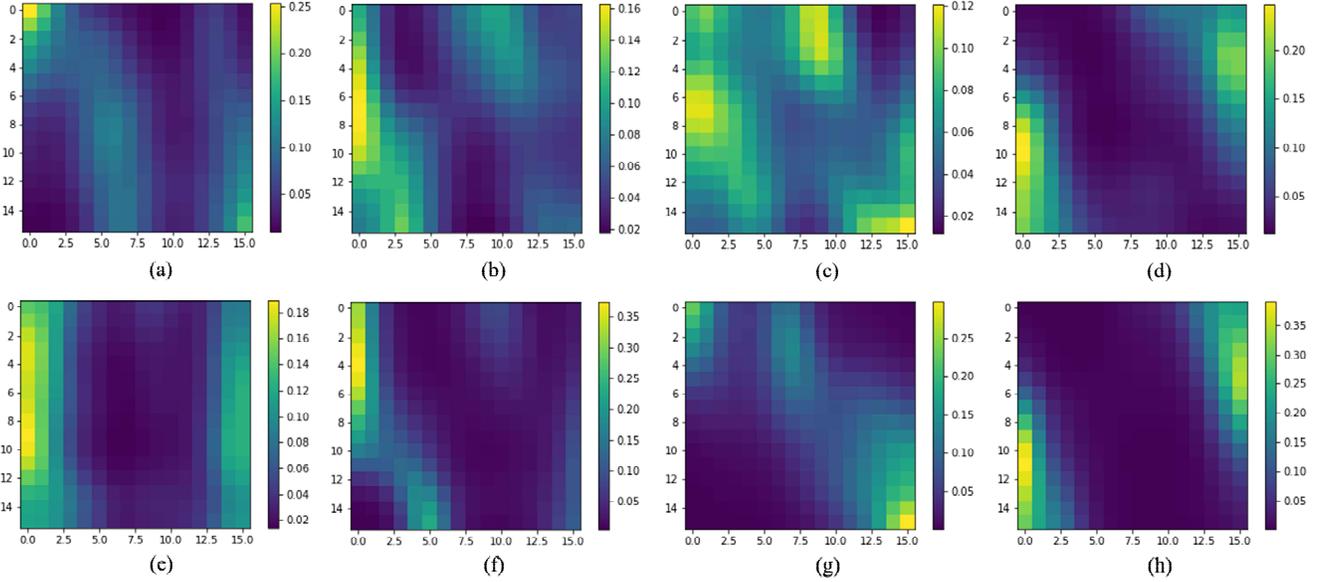}

\caption{Attention maps. Visualization of attention values among different attention heads. The x-axis and y-axis correspond to input and target frames respectively. We visualize the attention matrix value of the eight attention head of the first transformer encoder layer.(a) to (h) represent the attention value of eight attention heads separately. Lighter color indicates stronger attention value.}
\label{figure:attention_map}
\end{figure*}
\section{Experiments}
In this section, we first list the training/testing datasets and evaluation metrics we use for comparison. Then we compare our method with other state of the art frame-based and video-based methods. We also conduct multiple experiments to illustrate the influence of each component in our model. We provide both qualitative and quantitative results.
\subsection{Dataset}
As we all know, 2D datasets (annotated by 2D joint labels) contain a large amount of in-the-wild videos and their labels are easy to obtain. 3D datasets (annotated by 3D joint labels) contain less in-the-wild videos and their labels are more difficult to acquire, but can better supervise the 3D construction model than 2D ones. Therefore we conduct hybrid-training to leverage both 2D and 3D datasets. We use InstaVariety\cite{kanazawa2019learning}, Penn Action~\cite{zhang2013actemes} and PoseTrack~\cite{andriluka2018posetrack} as 2D datasets and 3DPW~\cite{von2018recovering}, MPI-INF-3DHP~\cite{mehta2017monocular} and Human3.6M~\cite{ionescu2013human3} as 3D datasets for training and evaluation.
Human3.6M is a large scale indoor dataset with 2D and 3D annotations. We use subjects S1, S5, S6, S7 and S8 as training data, and test our models on subjects S9 and S11.
3DPW is an outdoor dataset with 2D and 3D annotations. Following previous method, we also train our model on the 3DPW training set before evaluating on 3DPW test set. We also use AMASS~\cite{mahmood2019amass} dataset for adversarial training to discriminate a real/fake label for each sequence.

\subsection{Evaluation Metrics}
We use three image-based metrics: Mean per joint position error (MPJPE), Procrustes-aligned mean per joint position error (PA-MPJPE) and per vertex error
(PVE). Additionally, we report acceleration error~\cite{kanazawa2019learning}, which computes the average difference between the predicted and ground truth acceleration of each joint in $mm/s^2$. We use acceleration error as smooth indicator of temporal methods. A lower image-based metric error means a better performance in terms of accuracy of a model. And lower acceleration error means smoother results. 

\subsection{Comparison results}
As shown in Table~\ref{tab:aStrangeTable}, we compare our result with previous state-of-the-art methods. Our method outperforms VIBE in all image-based metrics on 3DPW and Human3.6M by a large margin. Besides, we achieve comparable result on MPI-INF-3DHP. Meanwhile, our result significantly outperforms all the image-based and video-based 3D reconstruction methods. The improvement of our model indicates that our two stream network helps to fuse spatial and temporal information, leading to more accurate and smoother 3D reconstruction. It also demonstrates that our transformer encoder and regressor are better at capturing temporal human motion information and predicting plausible SMPL parameters.\par

\begin{figure*}[t]
\centering
    \includegraphics[width=0.95\linewidth]{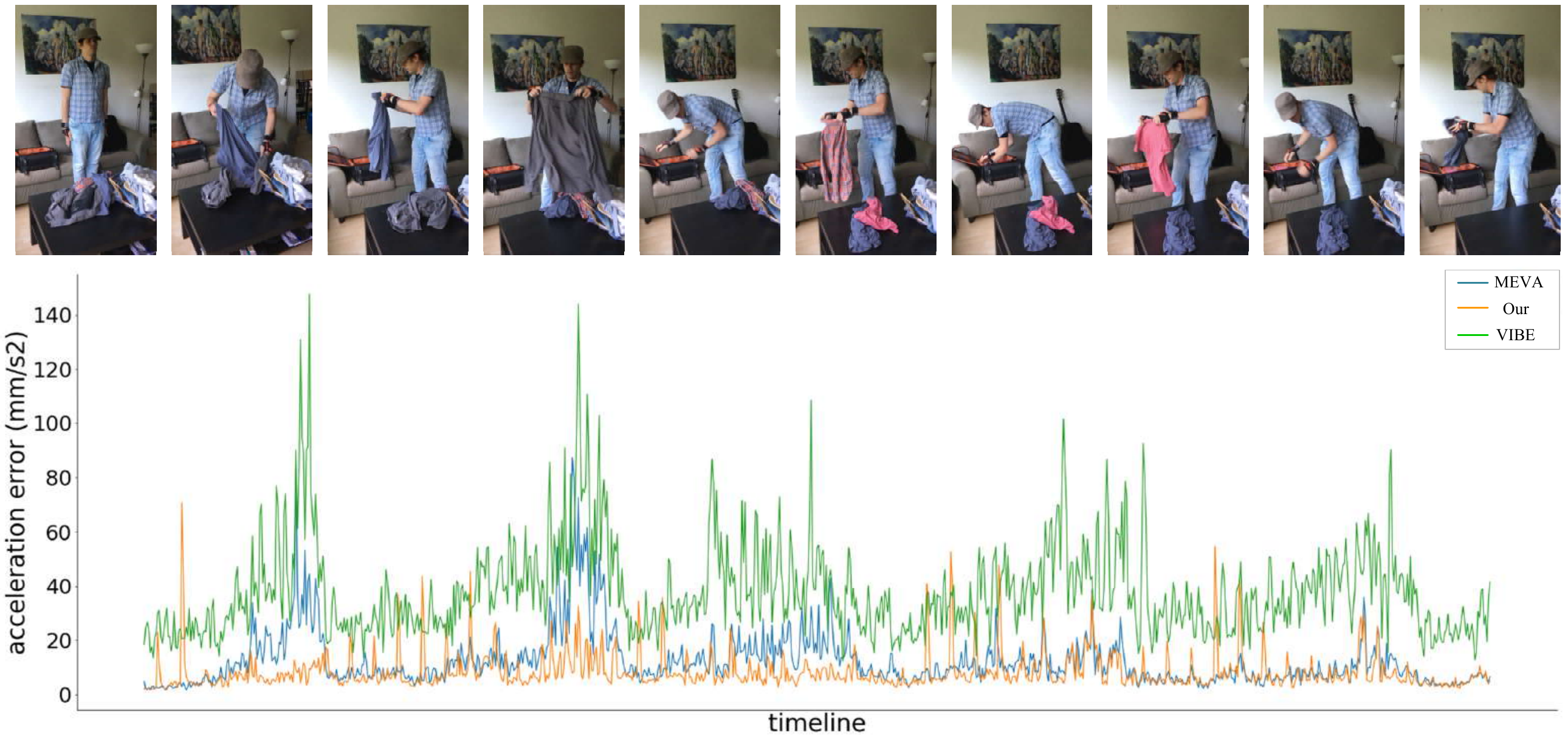}
\caption{\textbf{Comparison of acceleration errors between VIBE, MEVA and our method.} The top of the figure shows some samples from test images and the bottom graph is the acceleration error along the timeline for three models. Overall, our result is slightly smoother than MEVA and significantly better than VIBE.}
\label{figure:acc_err}
\end{figure*}

\begin{figure*}[t]
\centering
    \includegraphics[width=0.92\linewidth]{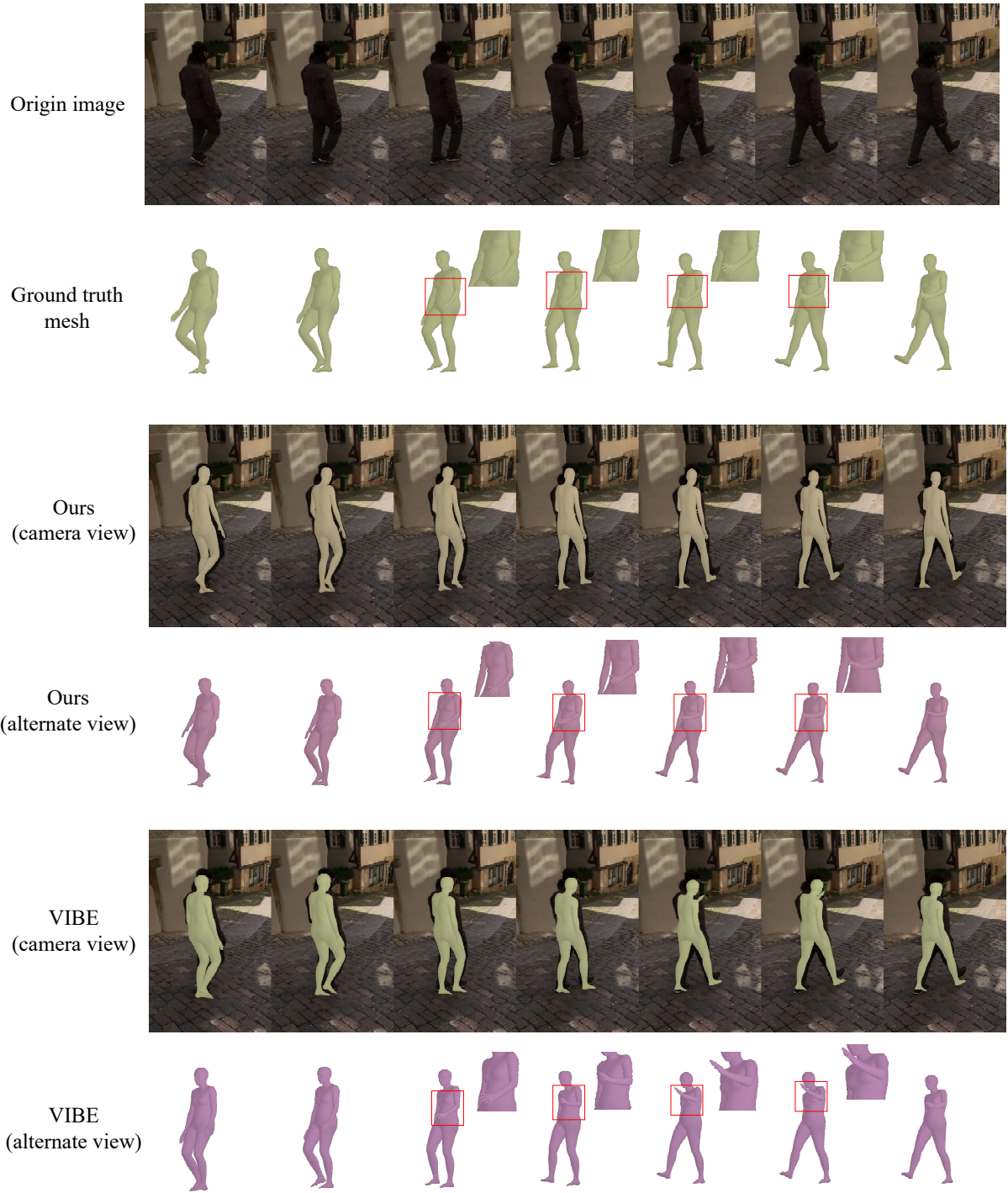}
\vspace{-2pt}
\caption{\textbf{Comparison visualization.} The first row corresponds to the origin video. The second row is the ground truth body mesh. The third and forth rows are our predicted results from camera view and alternate view. The fifth and sixth rows are VIBE results from camera view and alternate view.}
\label{comparison_visualization}

\vspace{-8pt}
\end{figure*}

Table~\ref{tab:aStrangeTable} shows the comparison on temporal smoothness between DTS-VIBE and state-of-the-arts video-based methods. We present some visualization results in Figure~\ref{comparison_visualization} which demonstrates that our model is better than VIBE when part of the body is self-occluded. The acceleration error of DTS-VIBE is lower than VIBE. As shown in Table~\ref{tab:aStrangeTable}, we produce smoother result than MEVA, which is state-of-the-art model on temporal smoothness, in terms of the acceleration error. We reduce the acceleration error by 67.3\% on Human3.6M, while maintaining the comparable result on 3DPW and MPI-INF-3DHP. Figure~\ref{figure:acc_err} shows the acceleration error comparison between other methods and ours measured on a sample of 3DPW test set. Compared with VIBE, our model keeps acceleration error at lower level. And our result is slightly better than MEVA except some rare spikes.
\begin{table}[t]
\centering
\setlength{\tabcolsep}{5mm}{
\begin{tabular}{c|c|c}
\toprule
model & PA-MPJPE & Accel\\
\midrule
GRU + TR      & 51.1 & 31.6\\
TE + HR       & 50.6 & 17.6\\
TE + TR(ours) & \textbf{50.3} & \textbf{11.0}\\
\bottomrule
\end{tabular}}
\caption{\textbf{Ablation study on transformer.} We replace our transformer encoder (TE) with GRU and replace Transformer Regressor with HMR regressor}
\label{tab:ablation_transformer}
\end{table}

\begin{table}[t]
\centering
\setlength{\tabcolsep}{5mm}{
\begin{tabular}{c|c|c}
\toprule
model & PA-MPJPE & Accel\\
\midrule
w/o flow      & 54.7 & 12.9\\
w/ flow       & 52.7 & 13.2\\
w/ flow + flow loss & \textbf{52.9} & \textbf{12.3}\\
\bottomrule
\end{tabular}}
\caption{\textbf{Ablation study on flow.} We conduct experiments that removes flow feature and flow loss from our model.}
\label{tab:ablation_flow}
\end{table}

\subsection{Ablation study}
Table~\ref{tab:ablation_flow} and Table~\ref{tab:ablation_transformer} show the ablation study results with and without each component. For each experiment, we use 3DPW, MPI-INF-3DHP and Human3.6M for training, and 3DPW for evaluation.

\textbf{Effectiveness of Transformer}. To single out the improvement introduced by our transformer based temporal encoder, we revert the temporal encoder into GRU and compare it with our model. Similarly, we also use regressor from ~\cite{kanazawa2018end} and our transformer-based regressor to prove the effectiveness of our transformer regressor. Table~\ref{tab:ablation_transformer} shows that the use of transformer as temporal encoder not only improves the accuracy of reconstruction but also provides smoother results. It indicates that transformer can better understand the complexity and variability of real human motion.
In addition, our transformer-based regressor significantly outperforms HMR regressor by 0.6\% on PA-MPJPE and 37.5\% on acceleration error. This demonstrates that our transformer regressor is able to predict more reasonable SMPL parameters.

\textbf{Effectiveness of optical flow}. We perform additional experiments to demonstrate the benefit of dense motion cues - optical flow. Here we remove human3.6M from training set. We also remove the flow-CNN feature and only use image feature to predict SMPL parameters. Table~\ref{tab:ablation_flow} shows the comparison between our model with and without flow feature. The accuracy of our model increases after adding optical flow features while maintaining high degree of smoothness.

\textbf{Effectiveness of flow loss}. We report the results with and without flow loss in Table~\ref{tab:ablation_flow}. It demonstrates that our acceleration error is reduced after using flow loss. Overall, the optical flow and flow loss together enhance the performance of model not only on the accuracy but also on the consistency. 

\textbf{Non-local Temporal Interaction.} To further understand the effectiveness of our DTS-VIBE model on temporal interaction, we visualize the attention map of transformer encoder. Non-local temporal information from input video are required to stabilize pose estimation while achieving high accuracy. Self attention mechanism of transformer can catch both long-term and short-term temporal information while GRU model from VIBE\cite{kocabas2020vibe} focuses more on catching short term temporal information.

Figure~\ref{figure:attention_map} shows the average attention value of eight attention head of the first transformer encoder layer. We sample 1000 video sequences from 3DPW test set and visualize its self-attention map. Each pixel indicates the self attention value of the target frame with respect to the input frame. Note that different attention head focuses on different input frames. For instance, attention head (d) and (h) tend to catch information from frames far apart while (c) and (g) tend to catch information within nearby frames.


 
\section{Conclusion}
In this paper, we introduce a two-stream 3D human reconstruction model named Deep Two-Stream Video Inference for Human Body Pose and Shape Estimation (DTS-VIBE), which consists of two-stream encoder, transformer-based regressor and motion discriminator. DTS-VIBE avoids extra inputs other than RGB. Additionally, we introduce a new flow loss to refine our model. Extensive experiments demonstrate the necessity and effectiveness of virtual multi-modality fusion in 3D human reconstruction. Furthermore, our model outperforms current state-of-the-art algorithms on 3D human reconstruction by a signiﬁcant margin.

\clearpage


\end{document}